\documentclass[10pt,twocolumn,letterpaper]{article}

\usepackage[pagenumbers]{iccv} 

\newcommand{\modelname}{SpoDify\xspace}

\usepackage{multirow} 
\usepackage{textcomp}
\usepackage[dvipsnames]{xcolor}

\definecolor{iccvblue}{rgb}{0.21,0.49,0.74}
\usepackage[pagebackref,breaklinks,colorlinks,allcolors=iccvblue]{hyperref}

\def\paperID{****} 
\def\confName{ICCV}
\def\confYear{2025}

\title{A Mesh Is Worth 512 Numbers: Spectral-domain Diffusion Modeling for High-dimension Shape Generation}

\author{Jiajie Fan\textsuperscript{1,2,\dag}\quad
Amal Trigui\textsuperscript{2,3,\dag}\quad
Andrea Bonfanti\textsuperscript{2}\quad
Felix Dietrich\textsuperscript{3}\quad
Thomas B\"ack\textsuperscript{1}\quad
Hao Wang\textsuperscript{1}\\
\textsuperscript{1} Leiden University, Niels Bohrweg 1, Leiden, The Netherlands\\
\textsuperscript{2} BMW Group, Bremer Str.~6, Munich, Germany\\
\textsuperscript{3} Technical University of Munich, Munich, Germany\\
\tt\small{\{j.fan,t.h.w.baeck,h.wang\}@liacs.leidenuniv.nl}\quad \\
\tt\small{\{babak.gholami, andrea.ba.bonfanti\}@bmw.de}\quad\\
\tt\small{\{amal.trigui, felix.dietrich\}@tum.de}}

\begin{document}
\twocolumn[{
\maketitle
\vspace{-37pt}
\begin{center}
    \centering
    \captionsetup{type=figure}
    \includegraphics[width=0.9\linewidth, trim=0mm 0mm 5mm 0mm, clip]{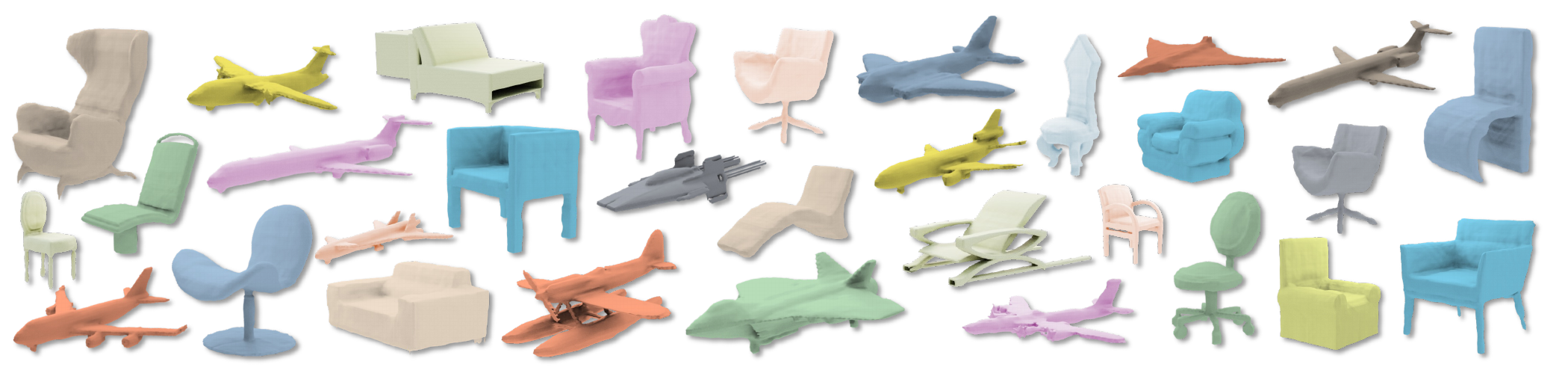}
    \captionof{figure}{A gallery of 3D meshes generated by \modelname.}
    \label{fig:Nurbs_generated_shapes}
\end{center}
 }]
\begin{abstract}
Recent advancements in learning latent codes derived from high-dimensional shapes have demonstrated impressive outcomes in 3D generative modeling. Traditionally, these approaches employ a trained autoencoder to acquire a continuous implicit representation of source shapes, which can be computationally expensive. This paper introduces a novel framework, spectral-domain diffusion for high-quality shape generation (\modelname), that utilizes singular value decomposition (SVD) for shape encoding. The resulting eigenvectors can be stored for subsequent decoding, while generative modeling is performed on the eigenfeatures. This approach efficiently encodes complex meshes into continuous implicit representations, such as encoding a 15k-vertex mesh to a 512-dimensional latent code without learning. Our method exhibits significant advantages in scenarios with limited samples or GPU resources. In mesh generation tasks, our approach produces high-quality shapes that are comparable to state-of-the-art methods.
\end{abstract}  
\footnote{\dag\ Joint first authors.}
\section{Introduction}
\label{sec:intro}
Generating 3D shapes represents one of the major challenges of the deep generative modeling community, where the researchers have made substantial progress in directly learning on mesh data~\cite{ranjan2018MeshAE, cheng2019meshgan, zhang2020meshingnet}. However, as a non-monotonous data representation, a mesh contains multiple modal data forms and their lengths vary with samples, where deep learning methods generally perform poorly. Most recently, research~\cite{chou2023diffusion, Shim_2023_SDF, hui2022neural} in this field enabled the generation of meshes with signed distance field (SDF)~\cite{stewart1993early}, a powerful implicit representation that encodes the source mesh into a voxel, where the value of each voxel grid indicates a distance value from the grid position to the nearest surface of the source mesh. Here, a negative value indicates that the point is inside the shape, while a positive value indicates that the point is outside the shape. Representing a mesh in the voxel form allows the implementation of 3D convolutional neural networks (CNNs), which addresses the challenge of using deep learning on meshes. However, to produce high-fidelity shapes, a large dimension of the voxel-shaped SDF is often required, e.g., $256^3$~\cite{hui2022neural, park2019deepsdf}, which poses computational and temporal challenges for directly learning with deep generative models (DGMs).

Meanwhile, the trend in high-dimensional data generation has shifted toward encoding the source data into a low-dimensional latent space so that DGMs can efficiently learn from compact latent codes. Using a trained autoencoder is the most commonly used methodology. It has yielded powerful DGMs, e.g., latent diffusion models (LDMs)~\cite{rombach2022ldm, zeng2022lion}, which can generate high-dimension data with much reduced computational cost. Encoding high-dimensional data has also been explored in the context of SDF representations~\cite{mittal2022autosdf, li2023diffusion, chen2019learning}. However, the quality of the results generated by such pipeline largely depends on the performance of the autoencoder introduced, which remains challenging and computationally intensive to train. In fact, the amount of training samples needed to properly train an autoencoder drastically increases with the dimensionality and diversity of the target data, which tends to be impossible for real-world design cases.

\begin{figure*}[t]
    \centering
    \includegraphics[width=1\linewidth, trim=0 0mm 20mm 0, clip]{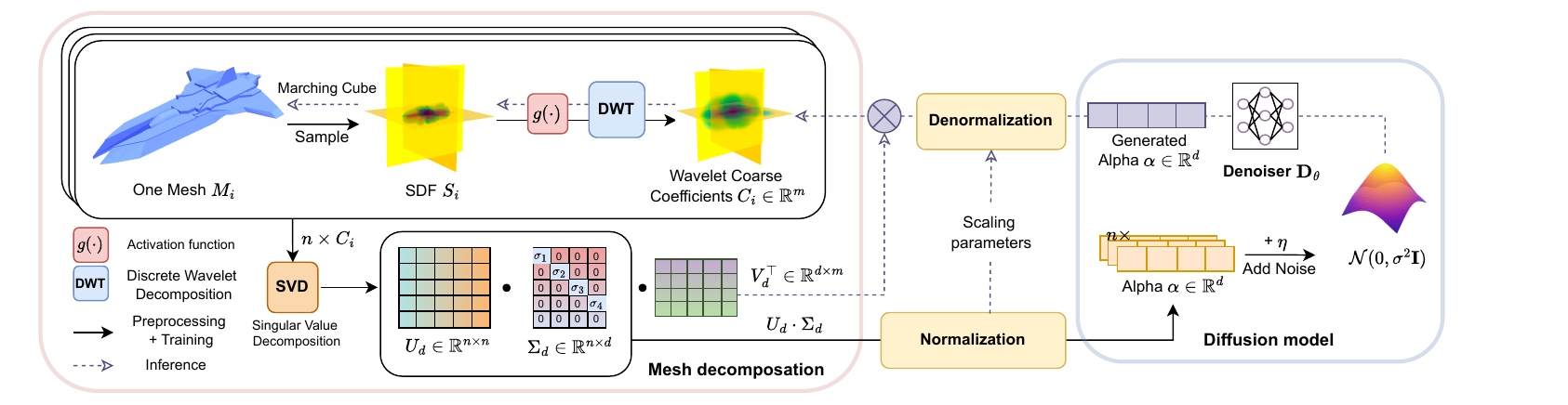}
    \caption{Diagram of \modelname. We apply singular value decomposition on a set of the coefficients that are derived by applying a signed distance field and discrete wavelet transformation on source meshes, resulting in the dataset of spectral features. Here, the basis $V_d^\top$ will be stored for later generation; spectral features $\alpha$ will serve as one sample and will be used for training the diffusion model. To generate a new mesh, the trained diffusion model generates new $\alpha$ for a given random noise. The generated $\alpha$ will be denormalized and then multiplied with pre-computed and stored $V^\top$ to obtain new low-frequency coefficients $C_i$, which can be converted to new mesh $M_i$.}
    \label{fig:Nurbs_GAT_Diagram}
\end{figure*}

Several existing approaches have used a learning-free encoding pipeline to obtain the latent variables of high-dimensional data~\cite{jones2004distance, canelhas2016eigenshapes, hui2022neural}. Among them, neural wavelet-domain diffusion~\cite{hui2022neural} (referred to as NWD in this paper) achieves state-of-the-art (SOTA) performance in generating complex topology and structures with clean surfaces and fine details. However, the introduced diffusion model has to be trained on 3D voxels of dimension 130$^3$ for a single-level wavelet transformation. To fill the gap between mesh generation and neural latent learning, a fully deterministic approach can be used, such as singular value decomposition (SVD), which has been historically used for dimensionality reduction in deep learning tasks including data classification~\cite{keegan2021tensor, yepes2024imageclassificationusingsingular} and image generation~\cite{kas2024eigengan}. \cite{kas2024eigengan} leverages SVD eigenvalues as a loss regularization term for GANs (Generative Adversarial Nets) training. Furthermore, SVD guarantees minimum information loss during the encoding, which is an essential characteristic for accurately reconstructing complex models. Even though SVD guarantees minimum information loss without the need for training, its application in generative 3d modeling remains understudied.

In this paper, we propose to exploit this idea and design a novel mesh generation method, spectral-domain diffusion for high-quality shape generation (\modelname), which uses a learning-free pipeline to encode meshes into low-dimensional spectral features that serve subsequently as the latent variables for training the diffusion-based DGM. We display our results in~\cref{fig:Nurbs_generated_shapes}, which are generated by learning on the ShapeNet dataset~\cite{chang2015shapenet}. Compared to SOTA methods (3DShape2VecSe~\cite{zhang20233dshape2vecset}, NWD~\cite{hui2022neural}) that rely on deep learning-based encoders or large data representations, our \modelname can produce comparable results, and in some cases even superior, by using generative modeling in a 512-dimensional spectral space.

\section{Related work}
\label{sec:relatedWork}

Huge progress has been made in using modern neural networks to tackle 3D generative tasks. It starts with representing 3D solids in voxels for its compatibility with convolutional mechanism, hereby yielding a set of early 3D generation models, e.g., 3D-GAN~\cite{wu2016learning} and 3D U-Net~\cite{cciccek20163dunet}. To avoid voxel's high memory and computation consumption, researchers have turned their attention to point clouds~\cite{qi2016pointnet, fan2017point, wang2019dynamic}. They represent shapes as a collection of discrete points in space, avoiding the large storage requirements of voxel grids. The generation of point clouds can be done by combining a point-cloud autoencoder and a GAN that is trained in the latent space of the autoencoder~\cite{achlioptas2018learning}; \citet{luo2021diffusion} enable the generation of high-quality 3D point clouds using diffusion models. In fact, point-based DGMs are extremely inefficient at modeling sparse data, and the lack of neighborhood information leads to poor model performance. Point-voxel CNN (PVCNN)~\cite{liu2019pvcnn} breaks the isolation wall between point-cloud and voxel. Building on that, the proposal of point-voxel diffusion (PVD) can significantly improve the fidelity of generated shapes. Moreover, LION~\cite{zeng2022lion} combines various advancements, such as PVCNN and latent diffusion model (LDM)~\cite{rombach2022ldm}, and achieves the novel state-of-the-art quality of generation results. Meanwhile, transformers are proposed for point cloud processing, e.g., \cite{guo2021pct, zhang20223dilg}. Nevertheless, point clouds lack connection information, which presents an important limitation in engineering applications, where surface continuity and structural integrity are crucial for activities such as modeling, finite element analysis, and production.

While point clouds fail to accurately represent the complex shapes and geometries encountered in industrial settings, meshes~\cite{lorensen1998marching, smith2006vertex} are the more commonly used form in industry and are the native representation of many CAE software and finite element tools. Nevertheless, due to the non-uniform representation of meshes (which consist of nodes and edges), it is nontrivial to leverage CNNs. To tackle this, MeshCNN~\cite{hanocka2019meshcnn} has been designed to enable using CNN on irregular mesh data forms and has achieved initial success in direct mesh learning. Afterward, substantial progress has been made in directly learning on meshes, e.g., convolutional mesh autoencoder (CoMA)~\cite{ranjan2018MeshAE}, MeshGAN~\cite{cheng2019meshgan} and MeshingNet~\cite{zhang2020meshingnet}. 

Most recently, researchers~\cite{chou2023diffusion, Shim_2023_SDF, hui2022neural} in this field have greatly facilitated the mesh generation task with the implicit representation of the signed distance field (SDF)~\cite{stewart1993early}. These methods are highly memory-efficient and are capable of producing smooth, high-resolution surfaces. Moreover, implicit functions are differentiable, which is an appealing property for ensuring smoothness over the latent representations, which are typically fed to state-of-the-art machine learning architectures. However, training models on implicit representations can often be computationally demanding and require large datasets. Conversely, MeshDiffusion~\cite{liu2023meshdiffusion} pioneered adopting an implicit representation form, utilizing Deep Marching Tetrahedra~\cite{shen2021deep} for shaping 3D objects. Another study~\cite{li2023diffusion} encoded local sections of 3D forms into voxel grids, training diffusion models within this framework. In an effort to simplify the learning process, SDF-Diffusion~\cite{shim2023diffusion} begins with a diffusion model on a coarse grid, followed by a patch-based enhancement technique to add intricate geometric particulars. Taking a different approach, LAS-Diffusion~\cite{zheng2023locally} adopts a gradual generation strategy, starting with a diffusion network for occupancy to create a basic sparse voxel grid before refining it to a denser resolution and then employing an SDF diffusion network to add localized details. Diffusion-SDF~\cite{chou2023diffusion} takes the process a step further by encoding shapes into triplane features and then compressing them into a concentrated latent vector, which the diffusion model uses to generate 3D shapes. Challenging the expressiveness of regular grids and global latent codes in geometric modeling, 3DShape2VecSet~\cite{zhang20233dshape2vecset} proposes a novel method of representing 3D shapes with a set of unevenly distributed latent vectors within the spatial domain. Most recent research in the unconditioned 3D shape generation~\cite{kalischek2022tetradiffusion} turns to a hybrid representation of explicit and implicit representations, namely a tetrahedral decomposition of 3D space, where 3D shapes are described via signed distances and deformation vectors on a tetrahedral grid with fixed topology.

Among methods that retain implicit representations, the closest work to ours is the method proposed by~\citet{hui2022neural} (we refer to it as NWD in this work), which transforms the 3D signed distance volume into wavelet coefficient volumes, enabling the diffusion model to construct an initial rough volume before a detailed predictor refines the geometric details. Also, \cite{zhou2024udiff} generates unsigned distance fields in the spatial-frequency domain with an optimal wavelet transformation. These approaches demonstrate how integrating implicit representations with wavelet decompositions can enhance the efficiency and effectiveness of generative modeling for 3D shapes.

\section{Method}
\label{sec:method}
Our method \modelname is inspired by NWD~\cite{hui2022neural}, where we additionally introduce an SVD-based decomposition approach to achieve a more interpretable and computationally efficient encoding of mesh representations. We show a diagram of \modelname in~\cref{fig:Nurbs_GAT_Diagram}.

\subsection{Spectral representation of mesh}
\paragraph{Clustering}
Our approach remains effective even in cases where only a limited number of training samples are available, as long as the utilized samples are representative of the larger dataset. We defer the explanation of this in~\cref{sec:mesh_generation}. To this end, we start by constructing a training set that ensures diversity while using fewer samples, for which we introduce a clustering process. To conduct the clustering, we first use diffusion maps~\cite{coifman2005geometric} on the complete set of available meshes, with the Chamfer distance as the metric. This embeds all available meshes into a shared diffusion space that preserves their geometric relationships. We then apply K-Means clustering to partition the dataset into $n$ clusters, capturing the diversity of the dataset. From each cluster, we select one representative mesh, ensuring that even with a small subset, the training set maintains a broad representation of the dataset’s variability. Further details on the clustering step and the diffusion maps representation are provided in the Supplementary Material.

\paragraph{Implicit representation with SDF}
Next, we represent the geometry of the mesh with the signed distance function (SDF) due to its differentiability and smoothness properties~\cite{foote1984regularity}. We sample from the $n$ representative meshes $M_{1,..,n}$. Each mesh $M_i$ is scaled to the range $[-0.5, 0.5]^3$ to standardize its size and position and then represented as an SDF $S_i$ of resolution $256^3$, so that 
\begin{equation}
\label{eq:sdf_formula}
    f_{M_i}(x) = \begin{cases} d(x , \partial M_i)\qquad & x\in M_i,\\
    -d(x , \partial M_i)\qquad & x\notin M_i,
    \end{cases}
\end{equation}
with $d$ being a suitable point-surface distance. When points in the field are far away from the shape surface, their value becomes large and unstable (i.e., with increasing deviation). These points are often identified as irrelevant by the model during training and contribute the least to the prediction of the final shape. To maintain smoothness and avoid discontinuities, \citet{hui2022neural} truncate the distance values in the signed distance function (SDF) to the range $[-0.1, 0.1]$. While this improves the learning, it does not emphasize accurate predictions along the shapes' contours. Unlike them, we introduce a limiting function

\begin{equation}
\label{eq:g_dot}
g(x) = \frac{1}{2}\tanh{(f_{M_i}(x))} - \frac{1}{2}.
\end{equation}
Through this bounding continuous function, SDF values far from the object's surface are constrained to approach zero. Since wavelets, applied in the following step, are inherently sensitive to local variations, this ensures that the resulting coefficients are more responsive to the shape boundaries rather than distant regions. As a result, with distant regions approaching zero, fewer wavelet coefficients are required to encode the surface, leading to a more efficient shape representation.



\paragraph{Pre-encoding with wavelet transformation}
Furthermore, we apply the 3D discrete wavelet transformation (DWT) on the preprocessed SDFs to extract localized features and patterns from the data. This step is essential for efficiently encoding localized spatial details while reducing redundancy in the representation. DWT can be considered as a particular type of convolutional layer with specific filter banks for extracting multi-scale features~\cite{oyallon2014genericdeepnetworkswavelet, guth2022waveletscorebasedgenerativemodeling}. Here, selecting an appropriate wavelet filter is crucial. While Haar wavelet is a popular choice for its simplicity, using it to encode smooth and continuous signals such as the SDF may introduce some voxelization artifacts~\cite{hui2022neural}. For the data representation chosen in this approach, the \texttt{Coiflet} wavelet~\cite{beylkin1991fast} is a suitable choice because, empirically, it provides a good balance between performance (in preserving important geometric features) and reconstruction accuracy. The application of the 3D DWT on each $S_i$ results in two sets of coefficients, i.e., one low-frequency coarse coefficient $C_i\in \mathbb{R}^{130^3}$ (referred to as DWT coefficients in this work) and high-frequency detail coefficients $D_i\in \mathbb{R}^{7\times 130^3}$.

\begin{figure*}[ht]
    \centering
    \begin{subfigure}[t]{0.49\columnwidth}
        \centering
        \includegraphics[width=0.8\linewidth, trim=75mm 20mm 65mm 60, clip]{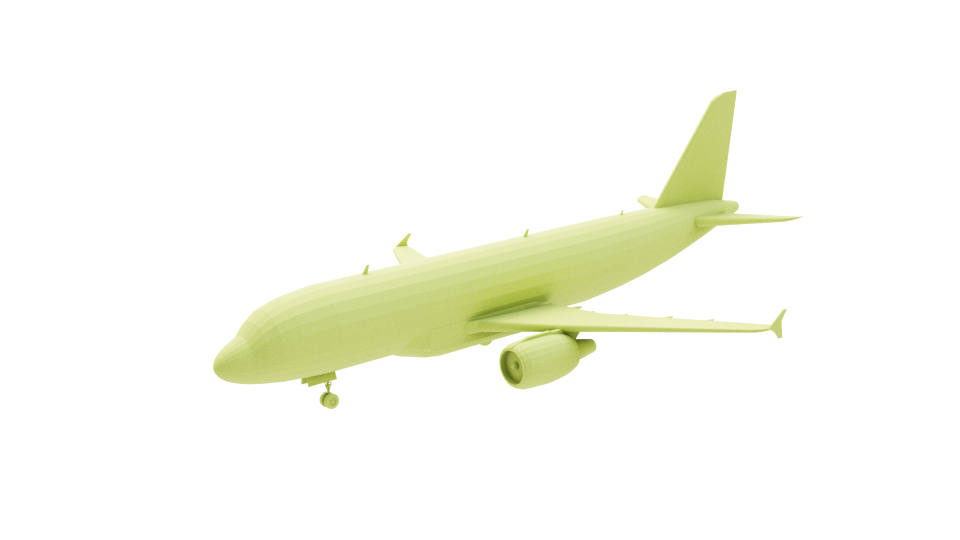}
        \caption{Source mesh}
        \label{fig:a}
    \end{subfigure}
    \hfill
    \begin{subfigure}[t]{0.49\columnwidth}
        \centering
        \includegraphics[width=0.8\linewidth, trim=80mm 20mm 65mm 60, clip]{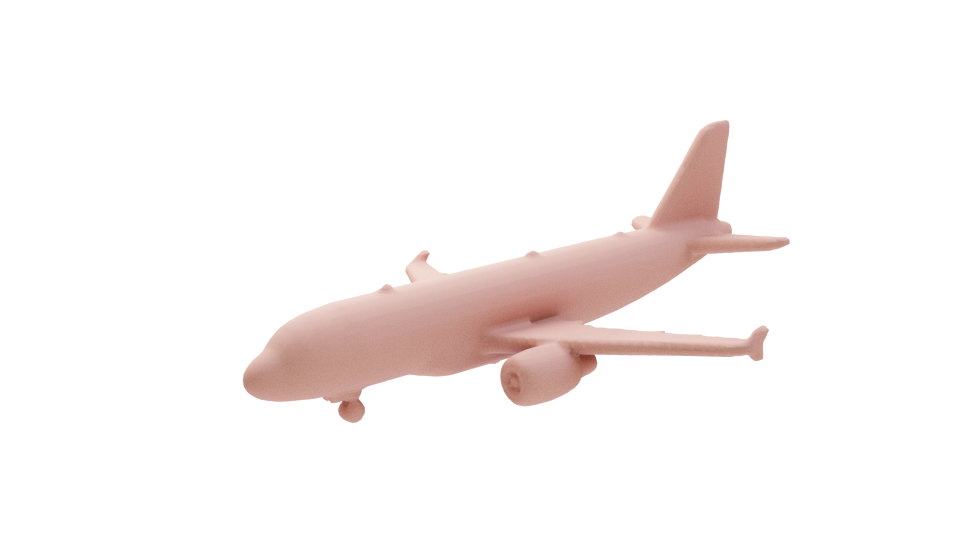}
        \caption{Detail and coarse coefficients}
        \label{fig:b}
    \end{subfigure}
    \hfill
    \begin{subfigure}[t]{0.49\columnwidth}
        \centering
        \includegraphics[width=0.8\linewidth, trim=80mm 20mm 65mm 60, clip]{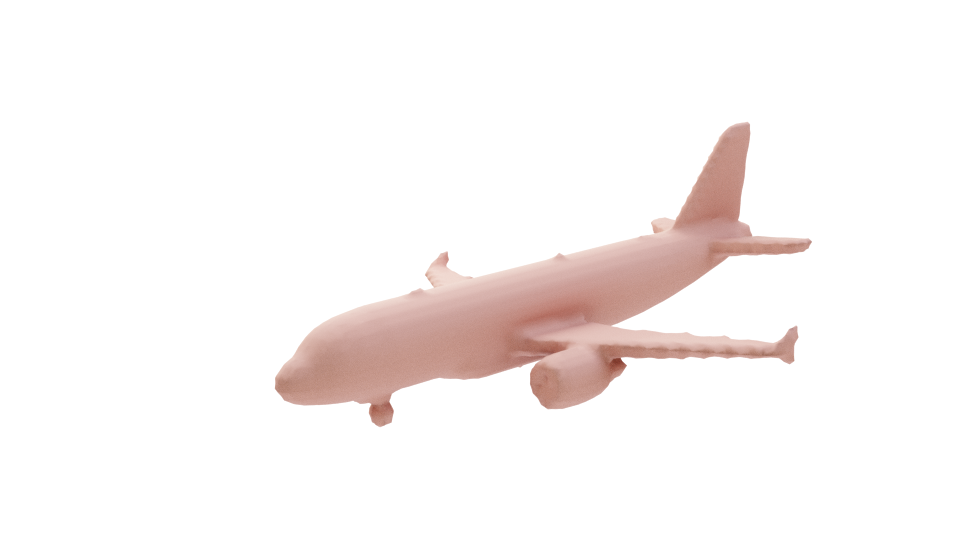}
        \caption{Coarse coefficients}
        \label{fig:c}
    \end{subfigure}
    \hfill
    \begin{subfigure}[t]{0.49\columnwidth}
        \centering
        \includegraphics[width=0.8\linewidth, trim=80mm 20mm 65mm 60, clip]{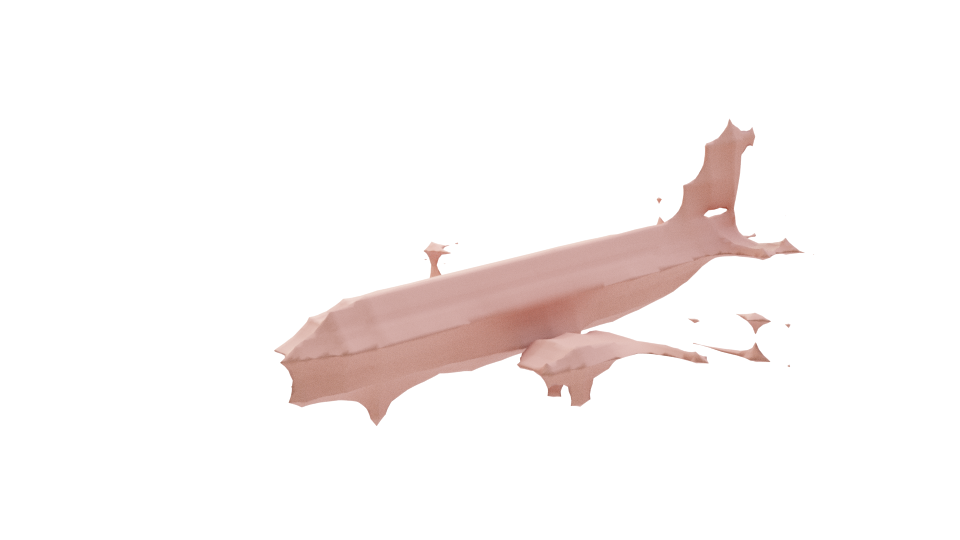}
        \caption{Coarse coefficients after 3-DWT}
        \label{fig:d}
    \end{subfigure}
    \caption{Effect of Wavelet Decomposition levels and the Dropping of High-Frequency Coefficients on Plane Mesh Reconstruction. (a) Original plane mesh; (b) Reconstructed plane after applying wavelet decomposition and reconstruction using all coefficients (both coarse coefficients and fine coefficients); (c) Reconstruction after one single-level wavelet decomposition level, keeping only low-frequency coefficients (coarse coefficients) and setting others to zero; (d) Reconstruction after \textbf{three} levels of wavelet decomposition, keeping only low-frequency coefficients (coarse coefficients) and setting others to zero.}
    \label{fig:coefficients_dropping}
\end{figure*}

For the subsequent process, we drop the detail coefficients, as a single-level wavelet decomposition retains sufficient information in the coarse coefficients for reconstruction. The difference between \cref{fig:b} and \cref{fig:c} demonstrates this effect. However, training a generative model directly on these coefficients ($C_i\in \mathbb{R}^{130^3}$) remains computationally demanding. To mitigate this, \citet{hui2022neural} applied hierarchical wavelet transformation for further compression. In such cases, discarding detail coefficients is no longer viable, shown in~\cref{fig:d}, as they must be further predicted from the coarse coefficients.

\paragraph{Dataset decomposition with SVD}
We propose to encode the DWT coefficients with SVD, which can be conducted through the following steps: (1) for $n$ meshes, flatten their DWT coefficients, denoted by $C_i\in\mathbb{R}^{m}, m=130^3, i=1,\ldots,n$, (2) stack the coefficients, resulting in a matrix $X=[C_1, C_2,...,C_n]\in \mathbb{R}^{n \times m}$, and (3) perform singular value decomposition (SVD) on $X$. Let $r\leq \min(m, n)$ denote the rank of $X$, the compact SVD is $X = U\Sigma V^\top$, 
where $U\in \mathbb{R}^{n\times r}$ and $V\in \mathbb{R}^{m\times r}$ are semi-unitary matrices and $\Sigma=\operatorname{diag}(\sigma_1, \sigma_2, \ldots, \sigma_r)$ contains the singular values on it diagonal. We arrange the singular values in descending order, i.e., $\sigma_1 \geq \sigma_2 \geq \cdots \geq \sigma_r > 0$, for later truncation. Here, we address: 
\begin{enumerate}
\item For the geometric information of each mesh (stored in rows of $X$), the rows of $U$ compress it drastically when $m \gg n \geq r$, e.g., high-resolution meshes are used on a small 3D shape dataset with sample number $n$.
\item SVD can be used to obtain a low-dimensional rank approximation of $X$ by keeping only the $d<r$ largest singular values, where the approximation error depends on the spectrum of $X$. Let $\widehat{X}_d = U_d \Sigma_d V_d^\top$, where $\Sigma_d = \operatorname{diag}(\sigma_1, \ldots, \sigma_d)$ and $U_d\in\mathbb{R}^{n\times d}, V_d\in\mathbb{R}^{m\times d}$ obtained from $U$ and $V$ by only keeping the first $d$ columns, respectively. We have the approximation error: $
\|X - \widehat{X}_d \|_{\text{F}}^2 = \sum_{i=d+1}^{r} \sigma_i^2$. SVD achieves the optimal approximation error by the Eckart–Young–Mirsky theorem~\cite{mirsky1960symmetric}. If the spectrum of $X$ decays rapidly, then we can safely truncate off a large fraction of singular values and keep the error small.
\end{enumerate} 
In practice, we maximize computational efficiency by decreasing the approximation rank $d$ to the lowest value, where the reconstructed mesh shows no visually recognizable error, and measured infinity error should be acceptable. See~\cref{fig:ablation_k_EVS} for an example of selecting $d$ for the airplane dataset in ShapeNet~\cite{chang2015shapenet}. Intuitively, each row of $U_d$ is the low-rank representation of a mesh shape, and its corresponding singular value reflects its frequency in the entire data set. Thereby, we decide to define the \emph{spectral features} of the mesh shape by scaling each row of $U_d$ with its singular value, i.e., rows of matrix $U_d\Sigma_d$. We shall denote by $\alpha$ a row of $U_d\Sigma_d$.
Also, the column space of \( V_d \) is a subspace of the original wavelet coefficients, serving as a ``dictionary'' or ``basis'' for representing the DWT data. Thus, we define $V_d$ as \emph{DWT basis} in this work. In this setup, each (flattened) DWT coefficient is approximated by
\[
\widehat{C}_i = \alpha V^\top_d,
\]
where $\widehat{C}_i = C_i$ iff. $d=r$. In the sequel, we shall train a generative model on the spectral feature $\alpha$, and a new shape can be created by sampling a new $\alpha$ from the model and reconstructing the DWT coefficients with the matrix $V_d$. Note that the space where $\alpha$ lies maintains the same smoothness of the SDF space, as only continuous functions are applied to those.

\subsection{Spectral-domain diffusion}
\label{sec:sepctral_diffusion}
\paragraph{Diffusion model architecture}
For the diffusion model, we adapt the Denoising Diffusion Probabilistic Model (DDPM) architecture proposed by~\citet{ho2020denoising}. However, we replace the 2D convolutional layers with fully connected dense layers to better handle the 1D sequence nature of the spectral features $\alpha$. This modification is motivated by the fact that the values in $\alpha$ are ordered according to the weights of the corresponding eigenvectors, and dense layers are better suited to capture global patterns in such structured data. 
For training and inference, we utilize a score-matching generative model, specifically the Elucidating Diffusion Model (EDM)~\cite{karras2022elucidating, fan2023noiseschedulinggeneratingplausible}, due to its fast sampling and efficient training capabilities. The diffusion model is trained to predict the spectral features $\alpha$ from noisy inputs, enabling the generation of new $\alpha$ values that can be used to reconstruct novel meshes.

\paragraph{Training objective}
The spectral features $\alpha$ are normalized to the scale $[-3, 3]$ and used to train the diffusion model. The diffusion model is trained using a composite loss function designed to ensure accurate prediction of the spectral features $\alpha$ while preserving the structural integrity of the generated shapes. The loss function is defined as
\begin{equation}
\label{eq:complete_loss}
L =  (1-\lambda) L_{\alpha} + \lambda L_{C},
\end{equation}
where $L_{\alpha}$ ensures the model accurately predicts the spectral features $\alpha$, and $L_{C}$ acts as a regularization term to incorporate the precomputed DWT basis $V^\top$. The individual loss terms are defined as
\begin{equation}
\label{eq:alpha_loss}
L_{\alpha} = \mathbb{E}_{\sigma, \alpha, \eta} \lVert D_{\theta}(\alpha + \eta; \sigma) - \alpha \rVert^2_2,
\end{equation}
\begin{equation}
\label{eq:coeffi_loss}
L_{C} = \mathbb{E}_{\sigma, \alpha, \eta} \lVert D_{\theta}((\alpha + \eta) \cdot V^\top; \sigma) - \alpha \cdot V^\top \rVert^2_2,
\end{equation}
where, $\ln(\sigma) \sim \mathcal{N}(P_{\text{mean}},\: P_{\text{std}}^2)$, $\eta \sim \mathcal{N}(0, \sigma^2\mathbf{I})$, $D_{\theta}$ is the implemented neural denoiser and $V^\top$ is the DWT basis obtained from SVD. While $V^\top$ does not participate directly in the training process, it serves as a critical multiplication factor for shape reconstruction. The regularization term $L_C$ ensures that the generated $\alpha$ values, when combined with $V^\top$, produce coherent and structurally valid low-frequency coarse coefficients, as demonstrated in~\cref{sec:ablation_study}. 

\subsection{Mesh generation}
\label{sec:mesh_generation}
Our proposed pipeline yields a set of basis $V^\top$ for storing shape elements, and features $\alpha$ that serve as ``weights'' that can be combined with the basis and form new shapes. Thus, at the beginning of our pipeline, we introduce clustering to maximize the shape elements obtained in the basis $V^\top$, ensuring that when the model generates new spectral features, they can be leveraged to explore the shape space of the larger dataset.

During generation, the trained diffusion model produces new $\alpha$ values from random noisy inputs. These generated $\alpha$ values are denormalized with pre-stored scaling parameters ($\alpha_{min}\in\mathbb{R}^{d}$ and $\alpha_{max}\in\mathbb{R}^{d}$ estimated from source $\alpha$ among all training samples) and multiplied with the pre-stored $V^\top$ to reconstruct the low-frequency coarse coefficients $C_i$. Finally, the inverse DWT and marching cube algorithms are applied to $C_i$ to generate a new mesh $M_i$.

\section{Experiments}
\label{sec:experiments}

\subsection{Experimental dataset and setup}
Evaluation is conducted on ShapeNet~\cite{chang2015shapenet} to compare with previous SOTA models. We focus on the airplane and chair categories from ShapeNet. These categories are chosen due to their geometric complexity and relevance in benchmarking generative models for 3D surface reconstruction. For mesh decomposition, we consistently select a sample size of $n = 1k$ for each dataset and set the reduced dimensionality to $d = 512$ as the default configuration. An experiment on tuning $d$ is presented in~\cref{sec:ablation_study}. Training is conducted on a single NVIDIA A10G GPU with a batch size of $32$ and a learning rate of $5\times 10^{-4}$ for $100k$ steps. Using the EDM training pipeline~\cite{karras2022elucidating}, we retain the original hyperparameters: $P_{\text{mean}}=-1.2$ and $P_{\text{std}}=1.2$. For inference, we set $\sigma_{\text{min}}=0.002$, $\sigma_{\text{max}}=80$, $\rho=5$, and $T=64$.

\begin{figure*}[ht]
\centering
   \begin{subfigure}[b]{\textwidth}
   \centering
   \includegraphics[width=\textwidth]{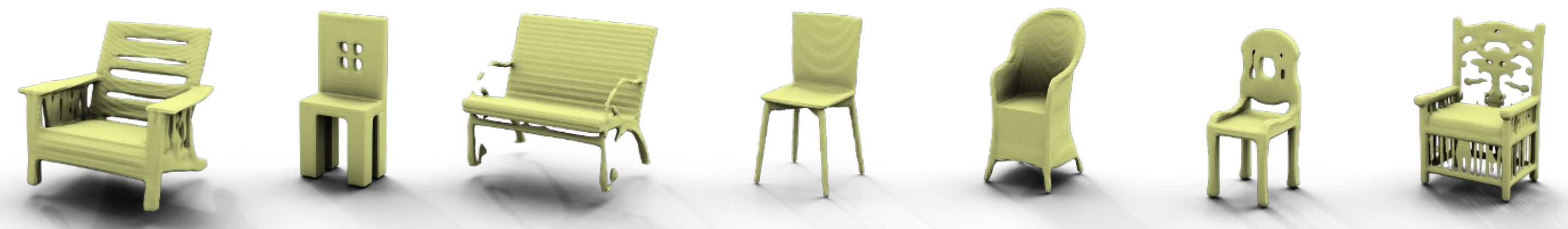}
   \caption{3DShape2VecSet \cite{zhang20233dshape2vecset}}
   \end{subfigure}
   \begin{subfigure}[b]{\textwidth}
   \centering
   \includegraphics[width=\textwidth]{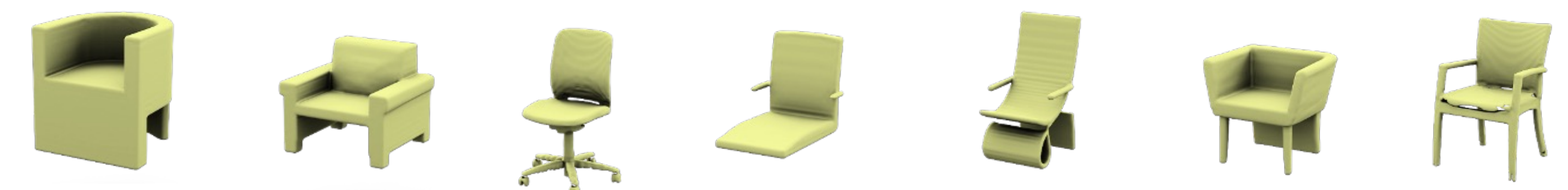}
   \caption{NWD~\cite{hui2022neural}}
   \end{subfigure}
   \begin{subfigure}[b]{\textwidth}
   \centering
   \includegraphics[width=\textwidth]{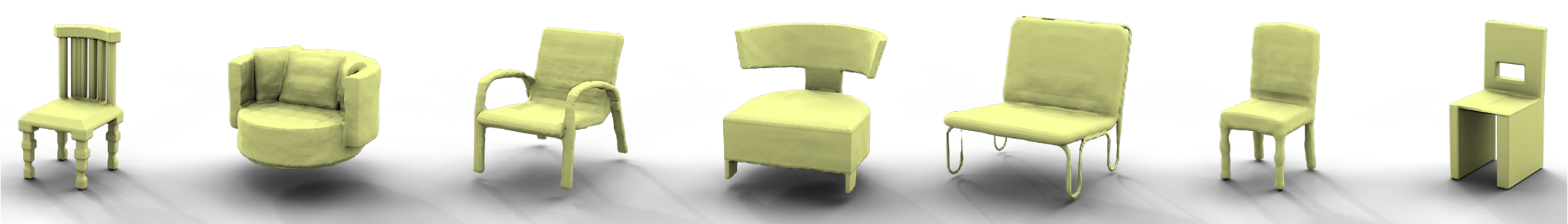}
   \caption{Our \modelname}
   \end{subfigure}
   
\caption{Qualitative comparison of chairs generated by different methods: (a) 3DShape2VecSet~\cite{zhang20233dshape2vecset}, (b) NWD~\cite{hui2022neural}, and (c) our \modelname.}
\label{fig:visual_SOTA_comp}
\end{figure*}

\subsection{Ablation study}
\label{sec:ablation_study}
\paragraph{Dimension of the spectral space}

\begin{figure}[ht]
\centering
\includegraphics[width=\columnwidth, trim=10mm 30mm 0mm 40mm, clip]{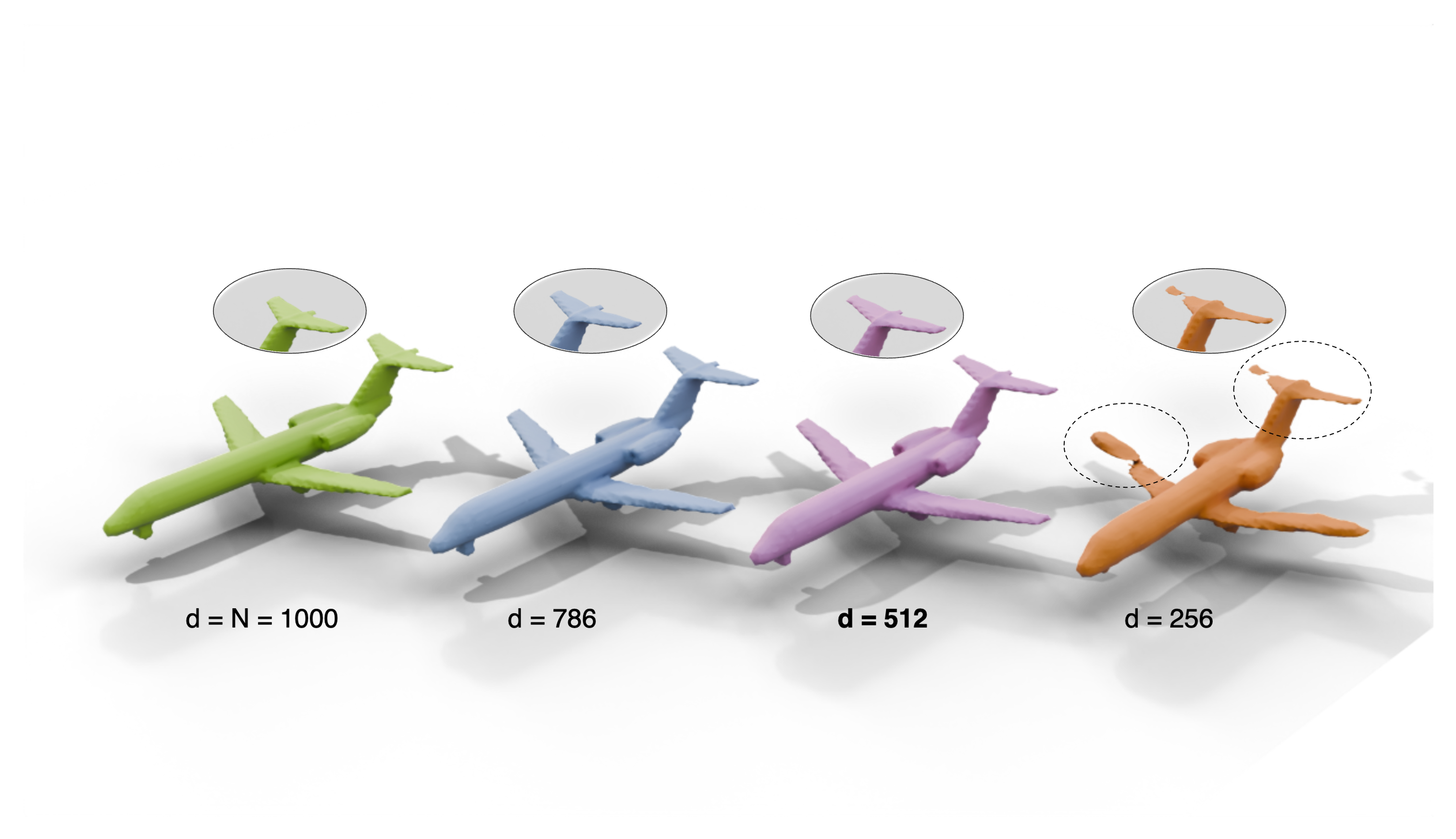}
\caption{Truncation level. Changing the reduced length $d$ of rows in $\alpha$ can impact the visual quality of final results and the computational power required to train the generative model. We notice that by truncating the row length until $d=512$, no significant visual artifacts are brought to the reconstructed meshes, whereas with $d=256$, reconstructed meshes show structural errors.}
\label{fig:ablation_k_EVS}
\end{figure}

Reducing the dimensionality of the spectral space $d$ directly contributes to reducing data size, model size, and computational costs at the expense of reconstruction accuracy. To investigate this trade-off, we conducted a hyperparameter tuning experiment, varying \( d \) across several values. The goal is to identify an optimal dimension that balances these competing factors while achieving strong overall performance. Based on the results, we select \( d=512 \) as the most suitable configuration for our method. \cref{fig:ablation_k_EVS} illustrates the effect of different truncation levels on various evaluation metrics, while~\cref{tab:abalation-k_EVs} provides a quantitative comparison of performance across various dimensions (\( d \leq n=1\,000\)). Metrics used for evaluation include minimum matching distance(MMD), Jensen-Shannon divergence(JSD), coverage(COV), and the \(L_2\)-norm reconstruction error.

At \( d=512 \), our method demonstrates a balanced performance across metrics, achieving the best trade-off between reconstruction accuracy and generative diversity. Specifically, the \( L_2 \)-norm reconstruction error improves significantly compared to \( d=256 \), dropping from \(1.54 \times 10^{-6}\) to \(5.82\times 10^{-7}\). While increasing \(d\) to 786 or 1\,000 further reduces reconstruction error, this comes at the cost of higher computational demands and a marginal decrease in generative diversity metrics such as JSD and COV. Dimension \( d=512 \) strikes an effective balance, delivering strong coverage (64.71\%) and reconstruction quality without unnecessarily increasing model size or computational overhead. 

\begin{table}[ht]
    \caption{Ablation study on the dimension of the spectral space out of $n=1\,000$ possible.}
    \label{tab:abalation-k_EVs}
    \small
    \centering
    \resizebox{0.8\linewidth}{!}
    {
        \begin{tabular}{c|cccc}
            \toprule
             $d$ &  MMD$\downarrow$ &  JSD$\downarrow$ &  COV$\uparrow$ &  L2 Recons$\downarrow$ \\
            \midrule
            256           & 1.68                     & 3.28                   & 50.06    &      1.54E-6         \\
            512         & 1.7           & 3.1                & 64.71      &  5.82E-7   \\
            786          & 1.81                    & 3.05                        & 69.55        &     1.49E-7     \\
            1\,000          & 1.76                   & 2.9                     & 61.05              &  0  \\
            \bottomrule
        \end{tabular}
    }
\end{table}
\paragraph{Training loss}
Several techniques have been introduced in our method, i.e., the loss regularization $L_C$ (~\cref{eq:coeffi_loss}) and the limiting function $g(\cdot)$ (~\cref{eq:g_dot}). Thus, we evaluate the impact of each feature through an ablation study. Here, we consider the following ablated models:

\begin{enumerate}[label=(\roman*)]
    \item \label{item:ablation_1} Ablation (w/o $L_C$): Removing the loss regularization of wavelet coefficients $L_{C}$;
    \item\label{item:ablation_2} Ablation (w/o $L_\alpha$): Removing the loss term of spectral feature $L_\alpha$;
  \item\label{item:ablation_3} Ablation (w/o $g(\cdot), L_C$): Deactivating the limiting function on SDF $g(\cdot)$ and removing the loss regularization of wavelet coefficients $L_{C}$;

\end{enumerate}

\begin{table}
    \caption{Ablation study on training and regularization losses with the airplane Dataset from ShapeNet. Configurations are formed by different combinations of parameters: $V^\top$ indicates the inclusion of regularization loss with respect to the wavelet domain, $g$ denotes the use of a limiting function on SDF values, and $\alpha$ specifies the inclusion of loss with respect to the spectral space.}

    \label{tab:abalation-tab1}
    \centering
    \resizebox{\columnwidth}{!}{
    \begin{tabular}{c|ccc|cccc}
        \toprule
              & \textbf{$g(\cdot)$} & \textbf{$L_C$} & \textbf{$L_{\alpha}$} & \textbf{MMD$\downarrow$} & \textbf{JSD$\downarrow$} & \textbf{COV$\uparrow$}  \\
        \midrule
        \modelname & \checkmark  & \checkmark & \checkmark  &  1.7 & \textbf{3.1} & \textbf{64.71}    \\
        Ablation \ref{item:ablation_1} &  \checkmark  &   &     \checkmark        & 1.776 & 3.27 & 61.05\\        
        Ablation \ref{item:ablation_2} &  \checkmark  & \checkmark  &  &           1.73 & 3.47 & 45.17    \\
        Ablation \ref{item:ablation_3} &     &        &    \checkmark   & \textbf{1.64} & 3.15 & 56.1 \\
        \bottomrule
    \end{tabular}
    }
    \vspace{-4mm}
\end{table}

\begin{figure}[ht]
\centering
\includegraphics[width=\columnwidth, trim=4mm 0mm 10mm 0mm, clip]{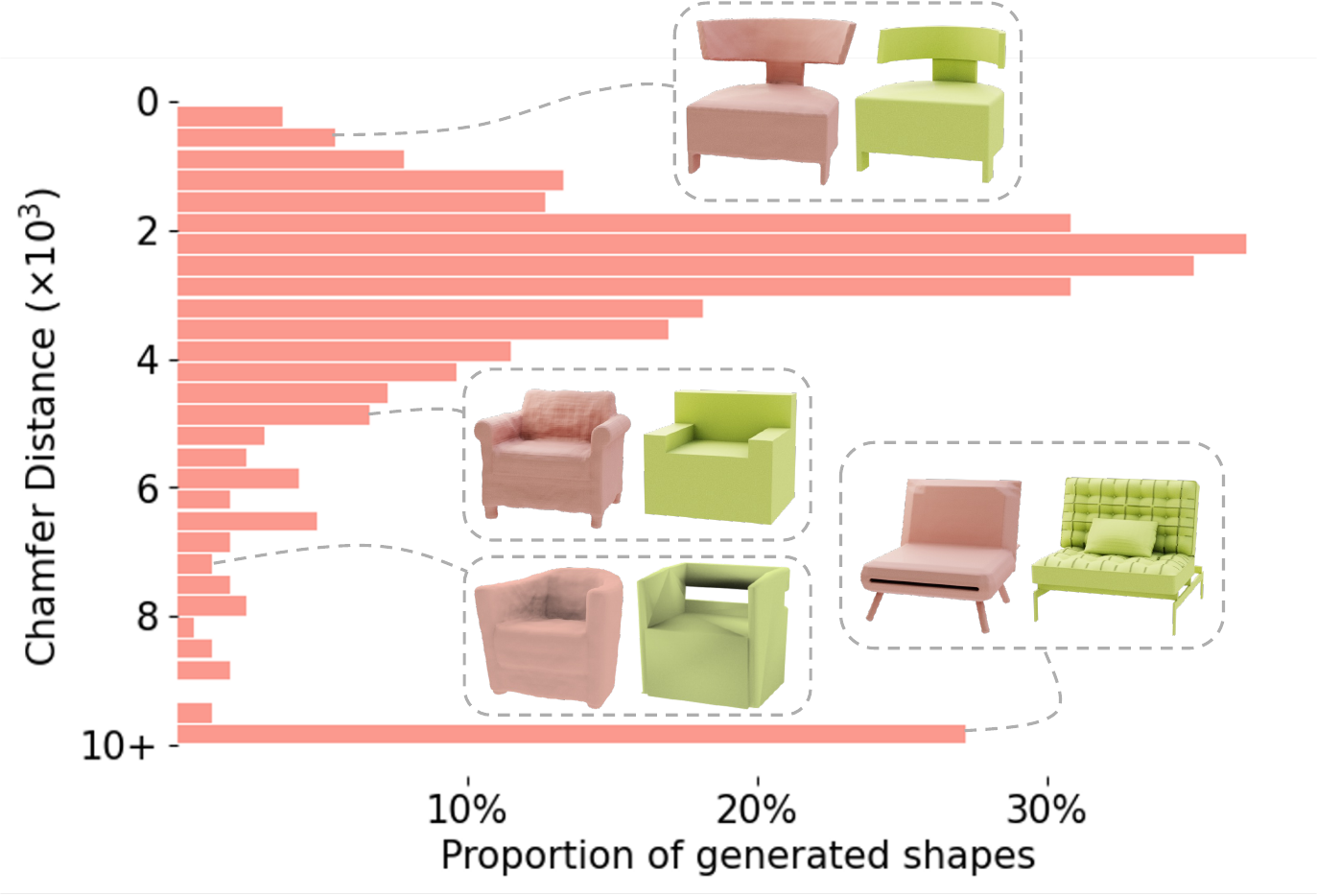}
\caption{Shape novelty analysis on ShapeNet~\cite{chang2015shapenet} chair category. We plot the distribution of 500 chair samples generated by our method and their closeness to the training dataset. Additionally, we display samples to visualize the similarity of various CD values, where \textcolor{YellowGreen}{\textit{green chairs}} are from the training dataset and \textcolor{RedOrange}{\textit{red chairs}} are generated.}
\label{fig:shape_novelty}
\end{figure}

\begin{table}[ht]
\caption{Quantitative evaluation of our proposed pipeline and current 3D shape generators. Metrics are computed over ShapeNet classes airplane and chair using the Chamfer Distance (CD). MMD is multiplied $\times 10^3$. NWD provided generated meshes for evaluation, while UDIff did not release its ShapeNet meshes. Due to computational constraints, we did not retrain UDiFF to compute JSD, resulting in missing values.}
\label{tab:1nna}
\centering
\setlength{\tabcolsep}{6pt}
\small
    \resizebox{\linewidth}{!}{
\begin{tabular}{@{}llcccccccc@{}}
\toprule
    \multicolumn{1}{@{}l}{Category} & Method & 1-NNA $\downarrow$ & MMD $\downarrow$ & COV $\uparrow$ & JSD $\downarrow$ \\
    \midrule
    \multirow{3}{*}{Airplane}
    & NWD \cite{hui2022neural}    &  97.24 &  1.69  &   71.2& 2.9   \\
    & UDiFF \cite{zhou2024udiff}   & 
    \textbf{74.48}  & \textbf{0.315}  & \textbf{64.77} &  -   \\
     & \textbf{ Ours}     & 97.98 &  1.7 &   64.71  &   3.1  \\ 
    \midrule
    
    \multirow{3}{*}{Chair} 
    & NWD \cite{hui2022neural} &53.4 &   1.180 &   45.19 &   0.027  \\
     & UDiFF \cite{zhou2024udiff}& 65.96 &  1.167  & 52.58  &  -   \\
     & \textbf{ Ours}      &   \textbf{46.69} &   \textbf{1.114}  &   \textbf{53.5}  &    \textbf{0.022}    \\ 
     
\bottomrule
\end{tabular}
}
\label{tab:quantitive_results}
\end{table}

\begin{table*}[ht]
    \small
    \caption{\textbf{Efficiency comparison}. Our approach can use less than 90\% GPU compared to TetraDiffusion \cite{kalischek2022tetradiffusion} and be trained in a few hours compared to days needed for NWD\cite{hui2022neural} and UDiff\cite{zhou2024udiff}. For NWD and UDiFF, the (+12) indicates additional GPU memory used for training the detail predictor network, separate from the main network responsible for global coefficient training.} 
    \label{table:efficiency}
    \centering
    \resizebox{0.85\linewidth}{!}{
    \begin{tabular}{l c c | c c c| c}  
    \toprule
        \multirow{2}{4em}{\textbf{Method}} & \multicolumn{2}{c}{\textbf{Representation}} & \multicolumn{3}{c}{\textbf{Training}} & \multicolumn{1}{c}{\textbf{Inference}} \\
        \cmidrule{2-7} 
        & Dimension & Compression rate  & GPU (GB) & Speed (it/s) & Duration (h) & Speed (s/obj) \\
        \midrule
        NWD \cite{hui2022neural} &$256^3$ & $\left({46}/{256}\right)^3 = 5.8$\textperthousand& 5.3 (+12) & 15.3 & 84 & 3.6 \\ 
        TetraDiffusion \cite{kalischek2022tetradiffusion} & $192^3$ & $\left({192}/{192}\right)^3 = 100\%$ & 78.2 & 0.3 & - & 33.3 \\ 
        UDiFF \cite{zhou2024udiff} & $256^3$ & $\left({46}/{256}\right)^3 = 5.8$\textperthousand & 4.5 (+12) & 4.89 & 84 & 3.4 \\ 
        \midrule
        \textbf{Our \modelname} & $256^3$ & ${512}/{256^3} = 0.03$\textperthousand & 7.2 & 100 & 3 & 3.3 \\ 
        \bottomrule
    \end{tabular}
    }
\end{table*}

\noindent From the results presented in \cref{tab:abalation-tab1}, we have the following observations: (1) The full model, which includes all three components (\(L_C\), \(L_\alpha\), and \(g(\cdot)\)), achieves in average the best performance; (2) Removing the wavelet coefficient regularization (\(L_C\)) results in a slight increase in MMD and JSD, as well as a drop in COV, indicating that \(L_C\) helps in improving coverage and reduces discrepancy; (3) Removing the spectral-feature regularization loss (\(L_\alpha\)) leads to a notable increase in JSD and a significant decrease in COV, confirming that \(L_\alpha\) is key to maintaining the diversity and quality of the generated shapes; (4) When both \(g(\cdot)\) and \(L_C\) are removed, the model performs better in terms of MMD compared to removing \(L_\alpha\) alone, but it still lagged behind the full model in JSD and COV. This suggests that while the limiting function \(g(\cdot)\) and the wavelet regularization term help in coverage, they do not fully compensate for the loss of spectral regularization.

In conclusion, our findings emphasize the importance of all three components in achieving the best performance. The wavelet coefficient regularization \(L_C\) and the limitation function \(g(\cdot)\) contribute to model stability and coverage, while the spectral regularization loss \(L_\alpha\) is essential for maintaining high-quality outputs and preventing overfitting. The full model, with all components, strikes the optimal balance between MMD, JSD, and COV, demonstrating the effectiveness of our design choices.

\subsection{Results}
\paragraph{Qualitative comparison}
To evaluate the performance of our method, \modelname, we first conduct a qualitative evaluation against several leading 3D mesh generation models while focusing later on the trade-off between performance and model complexity. We display the qualitative comparison in~\cref{fig:visual_SOTA_comp}. The evaluation is performed on the ShapeNet categories airplane and chair, where we compare the results of \modelname to those of NWD \cite{hui2022neural}, UDiFF \cite{zhou2024udiff}, and 3DShape2VecSet \cite{zhang20233dshape2vecset}. Our primary objective is not to surpass the SOTA generation models but to demonstrate that our method can achieve comparable or near-SOTA performance with a significantly less complex architecture and less computational power. While being able to fully capture the diversity of the dataset, our model maintains the structural properties of the underlying dataset. We notice that generated samples rarely present floating material or other structural artifacts that are physically incoherent in practice.

\paragraph{Quantitative comparison}
In the further quantitative comparison with SOTA methods, as results shown in~\cref{tab:quantitive_results}, our \modelname performs competitively across various metrics, including 1-nearest-neighbor accuracy (1-NNA), minimum matching distance (MMD), coverage (COV), and Jensen-Shannon divergence (JSD), indicating that it can generate high-quality 3D shapes similar to those produced by more complex models. In particular, our model outperforms the state of the art on the chair dataset. Moreover, COV values are consistently large, as the implicit representation adopted ensures the capability of reconstructing training samples.

\paragraph{Shape novelty analysis} 
This study examines the capacity of our proposed method to generate novel shapes with respect to the ones in the training dataset. To do so, we build on the work by~\cite{siddiqui2023meshgpt} and synthesize 500 chairs using \modelname. All shapes (generated and from the training set) are preprocessed through normalization into a unit cube, ensuring a standardized and equitable framework for comparison. We employ Chamfer Distance (CD) as the metric to quantify the similarity between shapes. We use CD to determine the sample in the training dataset that is the most similar to the generated chair. From the CD distribution shown in~\cref{fig:shape_novelty}, we observe that our model generates not only shapes that closely match those in the training set (low CD) but also produces realistic shapes that differ significantly from the training set shapes (high CD).

\paragraph{Efficiency comparison}
\cref{table:efficiency} highlights the comparative efficiency of various methods regarding training and inference requirements. Additionally, it reports the compression ratio between the input mesh and its corresponding representation used as input to the training model. Our proposed method demonstrates outstanding improvements in training speed and duration due to the impressive compression rate while requiring minimum GPU usage, while achieving comparable state-of-the-art results. Specifically, our method achieves the fastest training speed at 100 iterations per second (it/s), thanks to its streamlined model backbone, which significantly outpaces competing methods such as NWD~\cite{hui2022neural} (15.3 it/s) and UDiFF~\cite{zhou2024udiff} (4.89 it/s). This acceleration reduces the time required for large-scale training scenarios, making it feasible to handle extensive datasets without excessive computational cost. For inference, our method achieves a speed of 3.3 seconds per object, marginally surpassing UDiFF (3.4 s/obj) and vastly outperforming TetraDiffusion~\cite{kalischek2022tetradiffusion} (33.3 s/obj). 

Moreover, our model leverages an efficient data compression strategy, achieving a compression rate of ${512}/{256^3}=0.03$ \textperthousand, which contributes to reduced memory overhead during processing. Meanwhile, 3DShape2VecSet~\cite{zhang20233dshape2vecset}, while effective in certain scenarios, is hindered by the exceptionally large size of its data representation—approximately 300 GB—making it challenging to download and impractical to evaluate comprehensively under typical resource constraints.

\section{Conclusion}
\label{sec:conclusion}

Unlike prior works that rely on training to store shape information within the autoencoder parameters, our approach introduces a novel application of singular value decomposition to extract two components from the source data: (1) spectral features, a compact latent representation utilized to train the generative model, and (2) the basis, which provides foundational information for data reconstruction and is preserved during decomposition for reuse during the generation process. We are the first work to use the outputs of singular value decomposition as inputs for training generative models, we demonstrate that its outputs (spectral features and basis) can be effectively leveraged for generative modeling, enabling the synthesis of novel shapes. 

Furthermore, our \modelname not only maintains high-quality reconstruction but also achieves significant improvements in scalability, reducing training times from days to hours and compressing data dimensionality from gigabytes to megabytes. 
These advancements mark a crucial step toward making 3D generative models practical for handling high-dimensional data, especially when the amount of data is limited, and open up new avenues for research and applications.

\clearpage
\pagebreak

{
    \small
    \bibliographystyle{ieeenat_fullname}
    \bibliography{main}
}

\clearpage
\setcounter{page}{1}
\maketitlesupplementary
\appendix

\section{Evaluation metrics}
\label{sec:evaluation_metric_supp}
Evaluating the unconditional synthesis of 3D shapes presents a challenge due to the absence of direct ground truth correspondence. To address this, we use established metrics, consistent with previous works, which include: 
\begin{itemize}
\item \textit{Chamfer Distance (CD)}: A metric used to measure the similarity between two point clouds.
\item \textit{Minimum Matching Distance (MMD)}: Measures the mean CD between a sample in the test dataset and its closest sample in the generated ones. Lower values are better.
\item \textit{Coverage (COV)}: The percentage of test data with at least one match after assigning every generated sample to its closest test data based on CD. Higher values are better.
\item \textit{1-Nearest-Neighbor Accuracy (1-NNA)}: An optimal value of 50\% is ideal.
\item \textit{Jensen-Shannon Divergence (JSD)}: Measures the distribution distance between test and generated data after converting point clouds into discrete voxels.
\end{itemize}

These metrics ensure a comprehensive evaluation of the generated meshes in both the 3D space and visual quality. Since it involves various tasks in our work, we explain the metrics together in this section.

\section{Clustering}
We first computed pairwise Chamfer distances between all 3D meshes in the dataset to quantify shape similarity. The Chamfer distance is implemented using the method provided in the GitHub repository \url{https://github.com/fwilliams/point-cloud-utils}.
The resulting distance matrix, which captures geometric similarities between shapes, is then transformed into a similarity kernel matrix using an exponential function, enhancing the representation of relationships between the meshes. To reduce the high-dimensional representation while preserving the dataset's intrinsic geometric structure, we apply diffusion maps~\cite{coifman2005geometric}, extracting the top 64 eigenpairs to generate a low-dimensional embedding for each mesh. This embedding effectively captures the most significant modes of variation in the dataset. Subsequently, we perform clustering on the diffusion embeddings using the K-Means algorithm, identifying $n$ cluster centroids as representative shapes that encapsulate the dataset's diversity.

\section{Details and Global Coefficients}
In this section, we explore the effect of wavelet decomposition on the representation of 3D meshes. Initially, the data is represented as a \(256^3\) cubic grid using Signed Distance Functions. While this approach enables efficient representation of the object, it inevitably results in a small loss of information while reconstruction through marching cubes, as demonstrated in \cref{fig:b}.

Wavelet decomposition introduces a trade-off in this representation. By applying a single level of decomposition and retaining only the low-frequency coefficients (while setting the high-frequency ones to zero), the representation is reduced to a size of \(130^3\). Despite this reduction, a faithful object reconstruction can still be achieved, as shown in \cref{fig:c}. This demonstrates that a lower decomposition level can retain essential details of the object while reducing the overall data size.

However, when opting for multi-level decomposition, only retaining low-frequency coefficients further reduces the dimensionality, often resulting in a size of \(46^3\). At this point, the loss of high-frequency information becomes significant enough that recovering the original mesh is no longer possible, as shown in \cref{fig:d}. This highlights the need to learn those coefficients to recover fine details through a second model \cite{hui2022neural, zhou2024udiff}, emphasizing the trade-off between compression and fidelity in multi-level decomposition.

Ultimately, our chosen approach balances compression and fidelity, preserving essential details while maintaining an efficient object representation.  

\begin{figure*}[ht]
\centering
   \includegraphics[width=\linewidth, trim=0mm 0mm 10mm 0, clip]{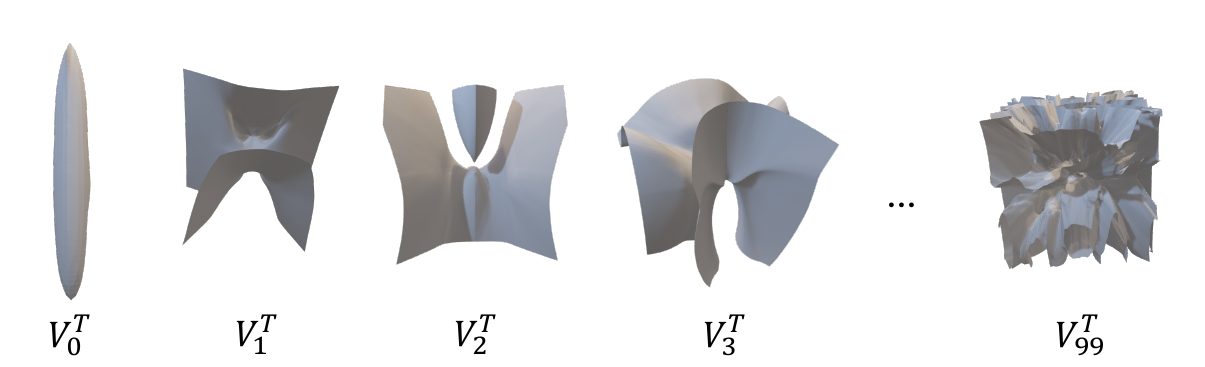}

\caption{Visualization of the DWT bases (the right eigenvector produced through SVD) of the ShapeNet Airplane dataset. Here, we only plot the first 4 DWT bases and the 99th DWT basis.}
\label{fig:Vis_EVs}
\end{figure*}

\section{Selecting approximation rank}
\label{sec:supp_selecting_rank_d}
\begin{figure}[ht] 
\centering
    \centering
    \includegraphics[width=\linewidth]{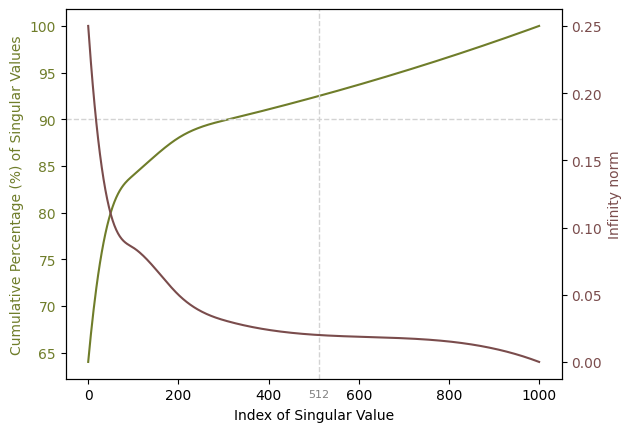} 
    \caption{Truncation level selection. Shown here is the trade-off effect of truncating the spectral feature rank $d$, analyzed by Infinity loss versus the cumulative percentage of singular values.}
    \label{fig:truncation_comparison}
\end{figure}

 In the following, we present the ablation study conducted on the airplane dataset. Based on the results shown in \cref{fig:truncation_comparison}, we selected \( d = 512 \) as the optimal configuration for our method. Furthermore, \cref{fig:truncation_comparison} supports this choice, demonstrating a high cumulative percentage of singular values (\>90\%) and a lower infinity norm. The infinity norm was calculated as the maximum absolute difference between the original SDFs and their reconstructions for each mesh, and then averaged across all shapes after truncating the spectral features, confirming the efficacy of the selected dimensionality.

\begin{figure}[ht]
\centering
\includegraphics[width=\linewidth]{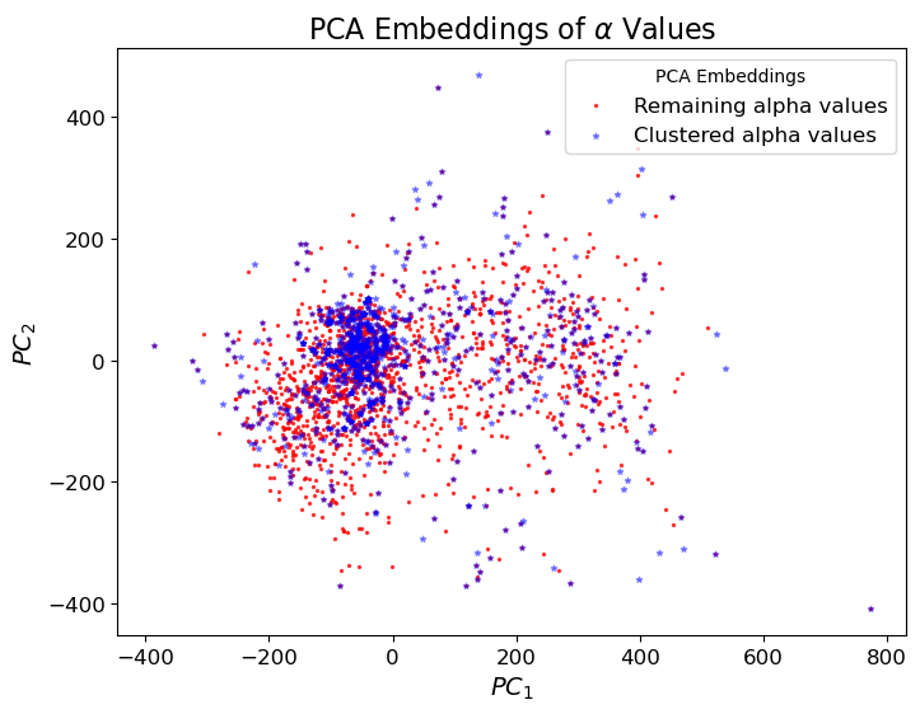} 
\caption{PCA Plot. The blue star markers represent the principal components derived from a set of clustered alpha values $\alpha = U \Sigma $. The red circle markers indicate additional data points ("remaining shapes of the dataset", $\alpha_{test} = X_{test} V $) projected into the PCA space.}
\label{fig:pca}
\end{figure}

\begin{figure*}[ht]
\centering
\includegraphics[width=\textwidth, trim=15mm 0mm 20mm 10, clip]{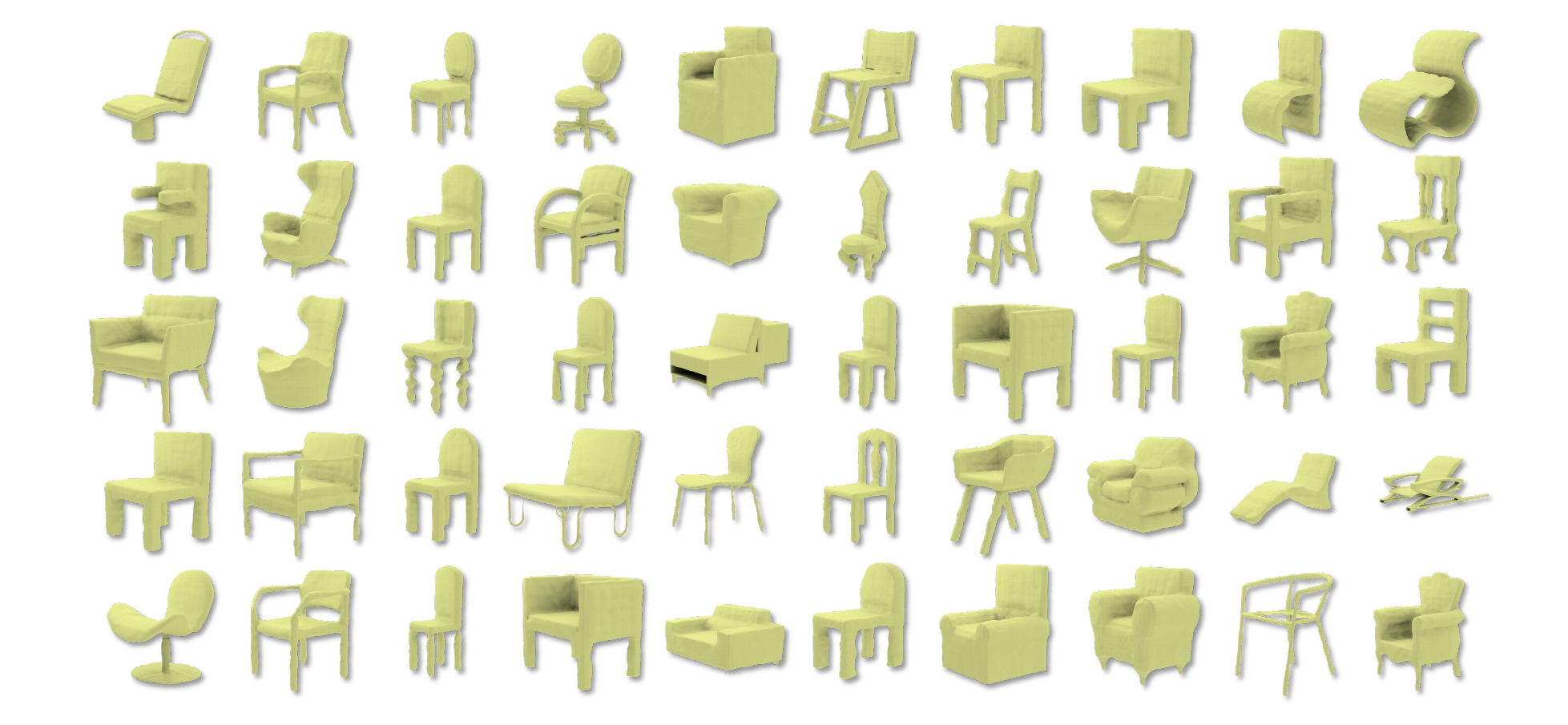} 
\caption{Examples of 3D meshes generated by our \modelname for qualitative evaluation. Extra examples can be found in the Supplementary Material.}
\label{fig:extra_chairs}
\end{figure*}

\begin{figure*}[ht]
\centering
   \includegraphics[width=\linewidth, trim=0mm 10mm 0mm 30, clip]{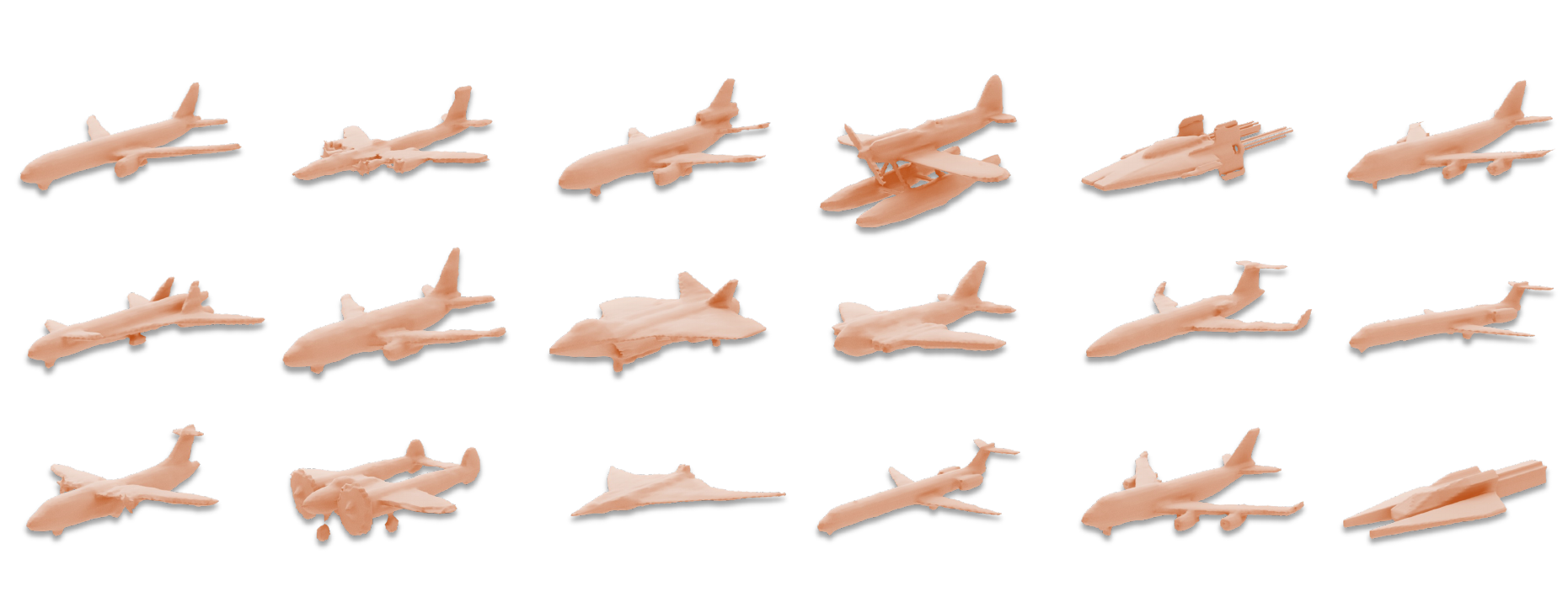}
\caption{Examples of 3D meshes generated by our \modelname for qualitative evaluation. Extra examples can be found in the Supplementary Material.}
\label{fig:generated_samples}
\end{figure*}

\section{Training Data}
The idea behind applying spectral decomposition, as previously motivated, is to represent an object as a combination of right eigenvectors in the eigenspace. These right eigenvectors act as the primary features that capture and assemble the dataset's structure, grouping and describing it in terms of these key characteristics. In the following \cref{fig:Vis_EVs}, we attempt to visualize the meaning of each eigenvector in the $V^T$
matrix. This is achieved by taking an eigenvector, applying the inverse discrete wavelet transform (IDWT), and then using marching cubes to map it back to the mesh domain. The result revealed distinct and interpretable patterns in the eigenvectors, with varying degrees of clarity depending on the dataset's complexity. In particular, the intricate and varied shapes of the airplane dataset presented challenges, as the lack of consistent features made it difficult to extract clear interpretations from the eigenvectors. Despite these challenges, the visualization process provided valuable insights into the underlying structures of the dataset.

Besides, the input data to the generative model are the eigenweights $\alpha = U \Sigma $. To analyze the training data, we employ dimensionality reduction techniques such as PCA and t-SNE to visualize the $\alpha$ data distribution. Additionally, we project non-clustered data points onto the eigenspace defined by the clusters to assess whether they align with the same distribution. In \cref{fig:pca}, we observe an alignment of the distributions. Training the model to capture this distribution is justified only if the unselected data resides within it, thereby confirming that the chosen clusters adequately represent the dataset's variability. This ensures that sampling or generating new points from the learned distribution remains meaningful when mapped back into the wavelet space.

\section{Limitation and future work}
Although we observed promising results with the airplanes and chairs from the ShapeNet dataset, our method has its limitations as well: (1) The dimension of the truncated spectral features should be proportional to the number of samples. This means that when applied to datasets with millions or even billions of samples, the dimensions of spectral features become correspondingly larger, thus affecting their efficiency; (2) using \modelname on multi-category might be challenging, as the current approach relies on a set of bases for storing the shape information. Extending this to multi-category datasets could present challenges, as the bases may not generalize well across diverse shape categories.

The journey of exploration extends beyond these results. There are several potential improvements to the current approach. An additional pipeline should be developed to incorporate the detail wavelet coefficients omitted in the encoding stage, which is essential for accurately generating smooth surfaces. Additionally, incorporating the wavelet coefficients in the training losses should be refined---one should extract and focus on the most crucial wavelets that have a decisive impact on the shape of the surface, thereby guiding the training process more efficiently. 

    
    
    
    
    

\end{document}